\documentclass[letterpaper]{article}
\usepackage{fullpage}

\usepackage[numbers]{natbib}
\usepackage[draft]{ktmacros}
\usepackage{algpseudocode}%

\newcommand{\dee}{\ensuremath{\mathcal{D}}}

\newcommand{\cee}{\ensuremath{\mathcal{C}}}

\newcommand{\err}{\ensuremath{\mathbf{e}}}
\newcommand{\calX}{\ensuremath{\mathcal{X}}}
\newcommand{\calY}{\ensuremath{\mathcal{Y}}}
\newcommand{\n}{\ensuremath{\mathbf{n}}}
\newcommand{\N}{\ensuremath{\mathbb{N}}}
\newcommand{\abar}{\ensuremath{\bar{a}}}
\newcommand{\Lauren}{Lauren}
\newcommand{\Evelyn}{Evelyn}

\title{Toward a Theory of Evaluability}
\author{Kunal Talwar\thanks{Apple. \texttt{ktalwar@apple.com}.}
}

\begin{document}
\maketitle
\begin{abstract}

Machine learning models, particularly large foundation models, are increasingly evaluated along multiple axes that are relevant for their usage. In addition to accuracy measures appropriate for the specific task of interest, measures capturing safety, calibration and fairness properties are often of interest. The classical approach to these problems assumes that the test set has been held out from the training process, an assumption that is increasingly harder to justify for large foundation models. In this work, we formalize a notion of evaluability without this assumption, and ask if it is feasible to evaluate black-box models in settings where our test samples may have been used during training.

We study this problem under different settings, capturing different levels of access to the model. We first study the natural setting, where the evaluator only looks at the output of the model on the test set. We fully characterize the class of distributions where accurate evaluation is possible, and show tight  upper and lower bounds on the sample complexity of evaluation. Our results show that black-box evaluation in this setup is feasible if and only if the distribution is close to being small support. We then consider evaluation algorithms that can query the model on additional inputs and demonstrate connections to self-correctors studied in program testing. Using these, we show that a small amount of query access can increase the power of the evaluator for natural distributions and concept classes.

\end{abstract}

\section{Introduction}

Imagine a learner Lauren releases a new model that gets $99\%$ classification accuracy on a public benchmark for a certain classification task. To determine whether the model is appropriate for use for a real-world use case, an evaluator Evelyn must get accurate estimates of the false positive and false negative rates of this model. How should Evelyn go about evaluating the model?

Existing approaches to estimation of such quantities rely on the availability of a fresh {\em test set} (sometimes the term {\em hold out set or holdout set} is used), with the assumption that this is a set of i.i.d. samples drawn from the distribution $D$ of interest, and that this set was not available to the training process that produced the model. This approach has been tremendously successful in pushing the state of the art and in the development of better ML methods.  Indeed decisions about practical ML models are typically based on performance on a {\bf fresh} test set, and the precise numerical confidence intervals in such situations are extremely important, as they help distinguish noise in the evaluation process from actual improvements in the models. When the test set is indeed a set of fresh, i.i.d. samples from $D$, this approach is mathematically sound and well-understood. Ensuring that the training process is not exposed to the test set is traditionally done by using separate training and test sets that are carefully curated (cf. related work on other contamination routes). However, with the increasing trend towards large models trained on all the data on the internet, curation may be infeasible. Indeed while Lauren may have taken care to ensure that the model training data does not directly include the benchmark test set, the test itself may be drawn from sources that are in the training corpus Lauren used to pre-train. A ``hidden'' test set may leak over time, either through intrusions or even through multiple measurements using it, and only large institutions may have the ability to maintain freshness and security. This approach of trusting the freshness of a test set is increasingly difficult to justify in broader settings.

An alternative approach that can get around this conundrum would be to use generalization bounds to control the gap between the accuracy (or another relevant measure) on the training set, and that on the distribution itself (cf. related work on generalization). However even relatively small neural network models have not yielded to numerous attempts to prove generalization bounds while preserving reasonable accuracy.
This approach is therefore unlikely to be effective for larger models in the near future.

Accuracy for an appropriate task is but one measure of interest. Various other important properties of ML models can be and are measured for determining their fitness for use in different contexts. Thus one is often interested in more refined measures such as false positive and false negative rates, confusion matrices, various measures of fairness, or safety, or measures of calibration and multi-calibration. Being able to accurately estimate any of these measures may often be a pre-requisite to deployment of such a model.

In this work, we initiate the study of evaluating statistical model properties without access to a fresh test set, or to the training process.
We first argue that score functions that are expectations of individual scores on examples are a reasonably broad class of score functions to evaluate. We show (\cref{sec:score_functions}) that this class includes several model properties that one may be interested in evaluating, and that other properties can be evaluated from these evaluations.

As a simple special case, let's consider a binary prediction task. We have a distribution $D$ over examples, labeled using a concept $c$. A learning algorithm $\mathcal{A}$ takes i.i.d. samples $(x, c(x))$ and outputs a predictor $f$. We measure the accuracy of the learning algorithm by the expected error between $f$ and $c$ on the distribution $D$. Classical results tell us that learning is infeasible without constraining (or making assumptions about) some of the objects of interest here. Distribution-independent learning theory leaves $D$ unconstrained, but may constrain $c$ and $f$. Generalization results alluded to above, which allow for provable evaluations, either constrain the predictor $f$, or constrain the algorithm $\mathcal{A}$ that produces $f$ in some way.

However, in the motivating settings above, constraining the predictor $f$ or even the training algorithm $\mathcal{A}$ in a reasonable way is infeasible. For example, the predictor $f$ may be a fine-tuned version of a large language model (LLM) and it is infeasible to know, let alone constrain the algorithm that produced the base model, or the data it was trained on. We will therefore assume in this work that both $f$ and $\mathcal{A}$ are unconstrained. In this setup, we study the question of evaluability of distribution and concept classes.

In a bit more detail, we assume we have access to a set of samples $\{(x_i,c(x_i))\}_{i=1}^n$ drawn from the distribution of interest, as well as the model prediction $f(x_i)$ on these examples. However, this sample may be available to the learning algorithm $\mathcal{A}$ as well. We characterize the class of distributions for which one can accurately evaluate a predictor $f$. Our results show that modulo trivialities, evaluability holds if and only if the distribution is close to being small support. Our result is quantitative and the number of samples needed to evaluate to accuracy $\alpha$ depends on the {\em amenability} parameters  of the distribution (formal definition in \cref{sec:prelims}) and on $\alpha$ in a simple quantitative way, that we show to be tight up to logarithmic factors.

Amenability is closely related to the distribution having a power law structure, where the head of the distribution captures most of the mass of the distribution. Real-world distributions are often amenable, and our results thus imply that for such distributions, one can indeed evaluate models without having any access to fresh samples. This has natural implications on the practice of evaluations: as an example, if we are evaluating a chatbot in a transit agency app, we must measure its performance on queries such as "How much is a monthly pass?" that users of the app are likely to ask of the chatbot. Performance on such queries is important even if this query occurs numerous times on the Internet, or in the model's training data. Attempting to avoid contamination by evaluating on a set of queries that excludes this query would be counter-productive in such a setting.%

However, there are many distributions of interest that are not amenable.  Our strong negative results motivate us to enrich the model. We enrich our evaluator by allowing it to get fresh unlabeled samples from $D$ and evaluate $f$ on new points. We show that self-correctors, introduced by \citet{BlumLR93}, can allow us to evaluate in this richer model. We give examples of natural function classes, where this allows us to do better for a large class of distributions.

Our results can be interpreted in a few different ways, and we present two different points of view next, starting with a pessimistic viewpoint. Our negative results suggest that effective evaluation is impossible without additional samples except when the input distribution is supported on a small set. Evaluation on stale benchmarks can be misleading and for faithful evaluation of properties of models, one must find a way to get fresh random draws from the distribution.

On the other hand, our results say that for amenable distributions, data contamination is not a concern. In some cases distributions of interest are indeed amenable. For example, \citet{ElondouBRGBMSHWK25} show that of real prompts issued to ChatGPT, more than 30\% fall into one of 66 ``tasks'', suggesting that a surprisingly small number of tasks cover the ``projection'' of a noticeable fraction of ChatGPT prompts. On a smaller scale, an analysis of the Reddit dataset~\citep{leaf} shows that in the collection of 4.2 billion word  occurrences with 3.46 million distinct words, 1,000 words account for nearly three quarters of word occurrences, 10,000 words account for nearly 93\% of all word occurrences, and 100,000 words cover nearly 99\% of all word occurrences. This suggests that, say, a text-to-speech model for English can be evaluated using a moderate number of words.
At a more abstract level, if one has a good embedding map, say from speech utterances or images to $\mathbb{R}^d$, then a real-world distribution may get mapped to an amenable distribution over clusters in this embedding space (e.g., embeddings of dog images may have a cluster for each dog breed). Indeed such assumptions have been made and evaluated in prior works \citep{ZhuAR14,Feldman20}.

Our results on evaluation with additional access suggest that it may be possible to effectively evaluate using additional unlabeled samples. Fresh unlabeled samples may often be easier to obtain compared to labeled ones, and if we can design efficient evaluators for tasks of interest, we could get around contamination issues and get accurate evaluations.  We remark that some existing approaches to evaluation, such as randomly permuting the ordering of options in multiple-choice questions~\citep{ZongYCZH24} are somewhat reminiscent of self-correction algorithms.

We view evaluation as related to, but different from benchmarking. Benchmarks have traditionally been used to measure and drive progress in machine learning. Even when they are not representative, or are overfit to, they have been shown to be reasonably good at comparing methods or models. Indeed \citet{RechtRSS19} show that image model performance on fresh samples is noticeably lower than that on existing benchmarks, but the ordering of models is largely preserved.
Evaluation, as we define it here, is concerned with the model's absolute performance on the distribution $D$ of interest.

\medskip\noindent{\bf Organization:} The rest of the paper is organized as follows. Additional related work is discussed in \cref{sec:related}. We then set up notations and some preliminaries in \cref{sec:prelims}. We define and give examples of our notion of score functions in \cref{sec:score_functions}, and formally define our notion of Evaluability in \cref{sec:defn}. \cref{sec:amenable_evaluable} shows that without any fresh data, evaluability is feasible if and only if the underlying distribution is amenable. We study a more powerful model in \cref{sec:oracle_access}, and we conclude with some open directions in \cref{sec:conclusions}.

\section{Related Work}
\label{sec:related}
The question of reliability of benchmarks that motivates this work has been raised as a concern in several previous works. A standard approach for a long time has been to have a separate ``hidden'' evaluation set, and while there have been instances of abuse of those~\citep{ilsvrc2015}, this approach largely worked in the pre-LLM era.

\citet{JacoviCGG23} argue that data contamination is an urgent crisis. They point out two routes to data contamination: training
data crawled from the internet, and training data
collected from calls to a closed API. Further, they point out that as larger models are often used to generate training data for smaller ones, it is increasingly hard to avoid contamination. They propose some mitigation strategies including encrypting test sets, and inserting canaries. \citet{ZongYCZH24} show that for multiple choice questions, randomly permuting the options led to significant decline in performance of models for several vision and language benchmarks. Other recent works~\citep{ZhangDLRWSZRZSKHKLY24,MirzadehASTBF25} have also presented compelling evidence of grade school math (GSM) benchmark performance not being an accurate measure of performance on fresh samples.

\citet{YangCZGS23} propose an LLM-based decontamination method, arguing that previously used n-gram-based methods often fail to prevent contamination.
\citet{ChandranSGSMM24} show how to build a cryptographic solution to this problem, by using a homomorphic encryption approach to evaluating a model on a hidden test set.
\citet{XieHZYCKKGJ24} study logical reasoning tasks of a certain kind, and use a programmatic approach to generate test data. This allows for an ``algorithmic'' benchmark.
\citet{FanHLLZ23} propose a benchmark based on combinatorial problems, with dynamic updates to avoid contamination.

It has been empirically observed that performance on possibly contaminated benchmarks may correlate well with performance on fresh samples. For image models, a beautiful line of work~\citep{RoelofsSRFHMS19,MillerKRS20,YadavB19,ShankarDRRRS21} has shown that there is a strong quantitative relationship between model performance on the original benchmarks (CIFAR10 and ImageNet) and a fresh test set generated using the same procedure. These results can be (and have been) interpreted as saying that re-using the same benchmark over decades has not led to overfitting to the test set and one would expect the same accuracy if one were to sample from the same distribution (with the drop in accuracy being attributed to distribution shift). With large language models trained on the whole internet, the contamination risk is a lot larger, and various techniques have been used to minimize the risk of contamination. While there is still correlation between performance on the ``stale'' benchmark and a fresh one (e.g. GSM8K vs GSM1K), this may not be sufficient to get accurate estimates of the model's performance on the distribution, which is the object of interest in the current work.

As alluded to earlier, there is a rich theory of generalization in machine learning that can give provable bounds when the model $f$ or the process generating it can be controlled. VC dimension bounds lead to {\em uniform convergence}, the phenomenon that all models in a concept class have empirical error close to the population error. This naturally leads to generalization bounds for every concept in the concept class. We refer the reader to~\citet{Shalev-Shwartz_Ben-David_2014} for a detailed treatment.
The algorithmic stability approach (e.g. \citet{BousquetE02}) to generalization argues that certain algorithms (namely those that satisfy a geometric stability property) generalize even though one may not have enough samples to guarantee uniform convergence over the class. A more recent line of work on adaptive data analysis, starting with~\citet{DworkFHPRR15} has shown generalization bounds under more relaxed information-theoretic notions of stability. More recently, \citet{GastparNSW24,GastparNSW24a} studied a question similar to ours but focused on generalization bounds, rather than error estimation. Recent concurrent work by~\citet{FeldmanKRTT26} studies a general data-dependent estimation question in a model where one can compute the event of interest on not just the given sample, but on other sets as well. In our motivating examples, this would correspond to being able to get access to multiple models trained, each trained on a specified training set.

Self-correctors were introduced in the work of \citet{BlumLR93} in the context of program testing, and by \citet{Lipton89}; and \citet{RubinfeldS96} first studied this notion for function classes. The linearity test introduced in the first work has been extremely influential in various parts of theoretical computer science, with the BLR test playing a major role in the PCP theorem. There is by now a long line of research on property testing~\citep{GoldreichGR98}, tolerant property testing~\citep{ParnasRR06} and self-correctors for various function classes.

Our work can be viewed as testing a certain property of the distribution of outputs of $f$. Property testing for distribution properties has a long and rich history. Recent work by \citet{HermanR22} proposed interactive proofs of distribution properties, and showed that in many cases, this can significantly reduce the sample complexity of the verifier.

\section{Notation and Preliminaries}
\label{sec:prelims}
We will denote by $\calX$ the (typically very large, or even infinite) data domain.
Given a (finite) label set $\calY$ we let $\Delta_{\calY}$ denote the set of all probability distributions on $\calY$, i.e. $\Delta_{\calY} = \{\mu : \calY \rightarrow [0,1] : \sum_y \mu(y) = 1\}$. When clear from context, we sometimes abuse notation and use $\calY$ to denote $|\calY|$.

We consider concepts $c : \calX \rightarrow \calY$. We say a distribution $D$ on $\calX \times \calY$ is consistent with $c$ if for all $(x,y)$ in the support of $D$, it is the case that $y = c(x)$. More generally, one may consider randomized label generation processes, whence $c : \calX \rightarrow \Delta_{\calY}$, and consistency says that for all $x$ in the support of $D_{\calX}$, the conditional distribution $Y|X=x$ is $c(x)$.   We will often consider a class of distributions and denote such a class by $\dee$.

A predictor is a function $f : \calX \rightarrow \Delta_{\calY}$, where $f(x)$ is viewed as the algorithm's guess for the distribution of $Y|X=x$.
Given two functions $f$ and $g$ on a domain $\calX$, and a distribution $D$, we define the distance $d_D(f,g)$ as $\E_{x \sim D}[d_{TV}(f(x),g(x))]$. As we have done here, when dealing with objects depending on $x$ alone, we will often abuse notation and say $x \sim D$, when we actually mean $(x,y) \sim D$.

We also recall some standard notions from learning theory.
\begin{definition}
\label{def:separates}
    Let $D$ be a distribution on $\calX$ and let $\cee$ be a concept class on $\calX$. For $\alpha >0$, we say that a set $S \subseteq \calX$ $\alpha$-separates $\cee$ w.r.t. $D$ if for any $c, c' \in \cee$, if $\Pr_{x \sim D}[c(x)\neq c'(x)] > \alpha$, then there is an $x \in S$ s.t. $c(x) \neq c'(x)$.
\end{definition}
\begin{definition}
    Let $\dee$ be a family of distributions on $\calX$ and $\cee$ a concept class on $\calX$. For $\alpha, \delta > 0$, the $(\alpha, \delta)$-separation complexity for $(\dee, \cee)$ is defined to be the smallest $n$ such that for any $D \in \dee$,
    \begin{align*}
        \Pr_{S \sim D^n}[S \mbox{ $\alpha$-separates $\cee$ w.r.t. $D$}] \geq 1-\delta.
    \end{align*}
\end{definition}
One of the fundamental results of learning theory is that for any $\dee, \cee$, the $(\alpha, \delta)$-separation complexity for $(\dee, \cee)$ is bounded by $O(\frac{VCDim(\cee)\ln \frac 1 \alpha}{ \alpha}+\frac{\ln \frac 1 \delta}{\alpha})$. We refer the reader to a standard textbook such as~\citep[Sec. 3.5]{KearnsV94} for a proof.

\subsection{Amenable Distributions}

We define a property of distributions that we call {\em amenability}, that captures the property that the distribution is largely concentrated on a small number of points.

\begin{definition}
Let $a_1,a_2,\ldots$ be a sequence of non-negative real numbers. A distribution $D$ on a domain $\calX$ is $\{a_t\}_{t\in \mathbb{N}}$-amenable if for all $t$, there is a finite set $U_t \subseteq \calX$ with $|U_t| \leq t$ such that $\Pr_{x \sim D}[x \not\in U_t] \leq a_t$. We say a distribution is amenable if it is $\{a_t\}_{t\in \mathbb{N}}$-amenable for some sequence $\{a_t\}$ with $\liminf a_t = 0$.
\end{definition}
A distribution with finite support is trivially amenable with the sequence $a_t$ that is 1 for $t \leq |Supp(D)|$, and $0$ otherwise. More precise bounds are possible when the distribution is a power law.
\begin{example}
Let $\calX = \{1,2,\ldots, K\}$ and for $\theta >0$, let $D$ be the distribution with $p_D(i) \propto  \frac 1 {i^{\theta}}$. This is often known as the {\em Zipfian} distribution. When $\theta >1$, this distribution is well defined even as $K \rightarrow \infty$, and is a {\em power law} distribution. It can be shown that $D$ is $\{\frac{1}{t^{\theta-1}}\}_{t\in \mathbb{N}}$-amenable when $\theta > 1$. For any fixed $K$ and $\theta>0$, one can derive a certain sequence for which the Zipf distribution on $[K]$ with parameter $\theta$ is amenable (though the analytic forms of this sequence  may not be particularly so). E.g. when $\theta=1$, the distribution is amenable with $a_t = 1-\frac{H_t}{H_K}$, where $H_m \approx \ln m + \gamma_{\mathrm{EM}} + \frac 1 {2m}$ is the $m$th Harmonic number; $\gamma_{\mathrm{EM}} \approx 0.5772$ is the Euler-Mascheroni constant.
\end{example}
\begin{nonexample}
    Let $\calX = [0,1]$ and let $D$ be the Lebesgue measure on $[0,1]$. Under this measure, any finite (or even countable) set has measure zero, and thus  $D$ is not amenable.
\end{nonexample}

Nominally, one can show that a probability measure is amenable if and only if it has countable support (i.e. is discrete). We believe that the amenability parameters of a distribution are important, and we will parameterize our results in terms of these parameters.
\section{Score Functions}
\label{sec:score_functions}
When evaluating a model, several properties of a model may be of interest. We will be interested in this work in properties that can be measured by reasonable access to a model $f$ and to a distribution $D$. To be concrete, we will focus on properties that are expectations of scores computable from individual examples. This can lead to different definitions, depending on the mode of access to $f(x) \in \Delta_\calY$. For some models, it is appropriate to assume that $f(x)$ itself can be computed. For example, a spam classification model may output a probability of an email being spam. For generative models (e.g. diffusion models) with large $\calY$, writing down $f(x)$ may be infeasible and one may only be able to sample from the distribution $f(x)$. For some other kinds of generative models (e.g. transformer models), one may also be able to query $f(x)$ at specific $y$'s of interest. In this work, we will focus on the first version with full access to $f$. This makes our negative results stronger. Our positive results can be extended to the weaker access models whenever the score itself is computable in the weaker access model.

We consider score functions that map a specific prediction $p \in \Delta_\calY$, and an example  $(x, y) \in \calX \times \calY$, and output a real-valued score. The score\footnote{While mathematically, a score can be viewed as a kind of a loss function, we choose to use a different term for a few different reasons. First, losses are typically associated with the parameters of the model, whereas in our settings of interest, only the model predictions are typically available. Secondly, the term loss carries a strong connotation of something to be minimized rather than measured, whereas our focus is on evaluation, not optimization. Finally, typically there is a single loss function one minimizes when training models. Fitness for use of a model typically depends on a host of different criteria, rather than on a single loss.} of a predictor $f$ on a distribution $D$ is then defined as
\begin{align*}
s(f; D) &\eqdef \E_{x \sim D, y \sim c(x)}[s(f(x); (x,y))].
\end{align*}

Many scores of interest depend solely on $f(x)$ and $y$ and not directly on $x$. In those cases, we will often abuse notation and write $s(f(x); y)$ instead of $s(f(x); (x, y))$.
In this work, we will always be interested in evaluating a claim that the score of a predictor is small. We give some examples.

\begin{example} [$0$-$1$ loss]
    Consider the score function $s(p;y) = 1-p(y)$. Then $s(f; D)$ measures the expected (in)accuracy of the classifier $f$ on the distribution $D$. This is the most commonly used measure of quality of a classifier.
\end{example}

\begin{example} [Squared loss]
    Suppose that $\calY = \{0,1\}$. Consider the score function $s(p;y) = (y-\E p)^2$. Then $s(f; D)$ measures the expected squared loss of the predictor that predicts the expectation of $f(x)$ on each example.

    More generally, when $\calY = \{1,2,\ldots,K\}$ is finite, we can define $e_y$ to be the $y$th basis vector in $\Re^K$.  Now consider the score function $s(p;y) = \|p - e_y\|_2^2$. Then $s(f; D)$ measures the expected squared loss.
\end{example}

\begin{example} [Log loss]
    Let $\calY$ be finite and consider the score function $s(p;y) = - \log p(y)$. This is the {\em cross-entropy} loss or the {\em negative log likelihood loss} that is commonly used to train models.
\end{example}

\begin{nonexample}[Expected Calibration Error]
Let $\calY = \{0,1\}$ and identify $\Delta_\calY$ with $[0,1]$ using the expectation map $p \rightarrow v(p) \eqdef p(1)$. For a classifier $f$, the Expected Calibration Error (ECE) is defined as $\E_{(x,y) \sim D}[|\E_{D}[y\mid f(x)] - f(x)|]$. This measure of calibration of a model is {\em not} a score according to our definition, as it cannot be estimated as an expectation over samples. In fact, \citet{BlasiokGHN23} point out that this measure has "fundamental theoretical flaws, which manifest as practical
frustration", including the fact that it is not estimable from samples.
\end{nonexample}

\begin{example} [Calibration Error Measures]
Let $\calY = \{0,1\}$ and identify $\Delta_\calY$ with $[0,1]$ using the expectation map $p \rightarrow v(p) \eqdef p(1)$. For a weight function $w : [0,1] \rightarrow \Re$, let $s_w(v;y) = w(v)(v-y)$ and let $s_w(f;D) = \E_{x \sim D, y \sim c(x)}[s_w(v(f(x)); y)]$. \citet{GopalanKSZ22} define a notion of weighted calibration error for a family $W$ as $s_W(f;D) = \sup_{w \in W} |s_w(f;D)|$. They show that several different measures of (mis)calibration of a classifier can be represented in this form for appropriate classes of weight functions $W$ with bounded Rademacher complexity. For example, interval calibration (which is quadratically related to the so-called upper and lower distances to calibration) can be expressed in this form where $W$ is the set of intervals in  $[0,1]$. Evaluating each $s_w$ with probability $(1-\delta)$ for small enough $\delta$ thus leads to evaluating $s_W$.
\end{example}

\begin{example} [Inaccuracy allowing Abstentions]\label{exa:abstain}
 Consider the setting where $\calY$ contains a special symbol $\star$ denoting abstaining to answer, and suppose that the concept class only contains functions $c$ whose range excludes $\star$. We can define a score function $s(p;y) = 1-p(y) - p(\star)$. Then $s(f; D)$ measures the expected (in)accuracy of the classifier $f$ on the distribution $D$, where the classifier is not penalized for abstaining.
\end{example}

\begin{example} [Accuracy Parity]
Let $G : \calX \rightarrow [0,1]$ be a predicate for membership in a group. Then the score function $s(p;(x,y)) = G(x)(1-p(y))$ can be used to measure accuracy within a group. Computing these scores for different groups can allow comparing accuracy of the model across groups.
\end{example}

\begin{example}[Hallucination Rate]
    This score is a variant of one that was considered in recent work by \citet{KalaiV24}. Suppose that for each $x \in \calX$, there is a small set $F_x \subseteq \calY$, and one can determine membership in $F_x$. For example, when a model generates a URL, we can check if the URL actually exists. Additionally $\calY$ may contain a $\star$ symbol corresponding to ``don't know''. Then $s(f(x), (x,y)) = \sum_j f(x)_j \one(j \not\in F_x \cup \{\star\})$ measures the incidence of hallucinated responses. Note that this can be estimated using samples from $f(x)$ alone.
\end{example}

We will assume all score functions in this work are bounded. While this is not quite true for some of our losses, e.g. log loss, one can often regularize the predictions to ensure that bounded-ness holds. In the rest of the paper, we will use the term {\em bounded score function} to mean one that satisfies this assumption.
\begin{assumption} (Bounded Scores)\label{assumption:bounded}
    For all $p \in \Delta_\calY$ and for all $y \in \calY$, $s(p; y) \in [0,1]$.
\end{assumption}

We state here a standard result on the estimation complexity of any score from fresh samples. We refer the reader to any textbook on concentration of measure (e.g.~\citep[Chap. 1]{Dubhashi_Panconesi_2009}) for a proof.
\begin{theorem}\label{thm:estimation}
    Let $D$ be a distribution on $\calX$, and let $s$ be a  bounded score function. Then for any $c$ the empirical estimate $\frac{1}{m} \sum_{(x,y) \in S} s(f(x), y)$ derived from a fresh random sample $S$ of size $m$ is $\alpha$-accurate if $m \geq \tau \cdot \frac{\min(s(f; D),1-s(f;D))\ln \frac 1 \delta}{\alpha^2}$, for a universal constant $\tau$. In particular, $m \geq  \frac{\tau\ln \frac 1 \delta}{\alpha^2}$ always suffices.
\end{theorem}

\section{Defining Evaluability}
\label{sec:defn}
We consider the following game defined by a distribution family $\dee$, a concept class $\cee$ and a score function $s$. Nature selects a $D \in \dee$ and a $c \in \cee$ such that $D$ is consistent with $c$.\footnote{In other words, we are looking at the realizable setting in this work. The definition can naturally be extended to the ``agnostic'' case where consistency with $c$ is relaxed to mean that $\Pr_{(x,y) \sim D}[y \neq c(x)]$ is small.} A sample of $n$ examples $S_n = \{(x_1,y_1),(x_2,y_2),\ldots,(x_n,y_n)\}$ is drawn i.i.d. from $D$ and provided to \Lauren\ and \Evelyn. \Lauren\ uses the $n$ samples, and potentially any other information she may have, to generate a predictor $f$. The $\sup_f$ (and $\inf_f$) in the definition below captures this potential dependency of $f$ on $S_n$. \Evelyn\ gets access to $S_n$, and to the evaluation of $f$ on each $x_i \in S_n$.
She must produce an estimate $E(S_n, \{f(x_i)\}_{i\in [n]})$ of the score $s(f; D)$.

\begin{definition} [Evaluability]\label{def:evaluability}
  Let \dee\ be a class of distributions, and \cee\ a concept class. Let $\err: \mathbb{N} \rightarrow [0,1]$. We say a score function $s(\cdot, \cdot)$ is $(\dee, \cee, \err)$-evaluable if there is an algorithm $E$ such that:
  \begin{itemize}
  \item {\bf Completeness}: For every $c \in \cee$, and every $D \in \dee$ consistent with $c$, it is the case that
  \begin{align*}
      \E_{S_n \sim D^n}[\sup_f \left(E(S_n, \{f(x_i)\}_{i\in [n]}) -  s(f;D)\right)] &\leq \err(n).
  \end{align*}
      \item {\bf Soundness}: For every $c \in\cee$, and every $D \in \dee$ consistent with $c$,  it is the case that
  \begin{align*}
      \E_{S_n \sim D^n}[\inf_f\left(E(S_n, \{f(x_i)\}_{i\in [n]}) - s(f;D)\right)] &\geq  - \err(n).
  \end{align*}
  \end{itemize}
  If $s$ is $(\dee,\cee,\err)$-evaluable for some $\err$ with $\lim_{n\rightarrow \infty}\err(n)=0$, we say that it is $(\dee,\cee)$-evaluable. In this case, we say that the evaluation complexity of $s$ for $(\dee,\cee)$ is $\n: [0,1] \rightarrow \N$ where $\n(\alpha) = \min \{n : \err(n) \leq \alpha\}$.
\end{definition}

In words, soundness says that up to the error $\err(n)$, \Evelyn's estimate of the score is a valid upper bound on the actual score of any predictor (even one that depends on the sample $S_n$). On the other hand completeness says that \Evelyn's estimate is (up to a small error) not larger than the true score. Together, these imply that the estimate is accurate up to an error of $\err(n)$. Separating the two requirements allows us to consider the following stronger definition.

\begin{definition} [Strong Evaluability]
  Let \dee\ be a class of distributions, and \cee\ a concept class. Let $\err: \mathbb{N} \rightarrow [0,1]$. We say a score function $s(\cdot, \cdot)$ is $(\dee, \cee, \err)$-strongly evaluable if there is an algorithm $E$ such that:
  \begin{itemize}
  \item {\bf Completeness}: For every $c \in \cee$, and every $D \in \dee$ consistent with $c$, it is the case that
  \begin{align*}
      \E_{S_n \sim D^n}[\sup_f \left(E(S_n, \{f(x_i)\}_{i\in [n]}) -  s(f;D)\right)] &\leq \err(n).
  \end{align*}
      \item {\bf Soundness}: For every distribution $D$, it is the case that
  \begin{align*}
       \E_{S_n \sim D^n}[\inf_f\left(E(S_n, \{f(x_i)\}_{i\in [n]}) - s(f;D)\right)] &\geq  - \err(n).
  \end{align*}
  \end{itemize}
\end{definition}

In words, strong evaluability says that the evaluator's estimates are {\em always} valid upper bounds on the score {\em without} any assumptions on the data distribution: the soundness holds for every $D$, even those inconsistent with any concept. If the data distribution assumptions hold, then they are also useful. Thus if a strong evaluator tells us that the model is accurate, we can be assured that the model is indeed accurate. In contrast, if a weak evaluator tells us that the model is accurate, we can be sure that {\em either} the model is accurate {\em or} the assumption that the data distribution $D$ is in $\dee$ is violated.

The following property of the definition will be useful.
\begin{observation}
    Suppose that $\dee \subseteq \dee'$, $\cee \subseteq \cee'$ and $\err(n) \geq \err'(n)$ for all $n$. If a score function $s$ is $(\dee', \cee', \err')$-evaluable, then it is also $(\dee, \cee, \err)$-evaluable.
\end{observation}

Looking ahead, in \cref{sec:oracle_access}, we look at settings where \Evelyn\ has additional power, e.g. oracle access to $f$, or access to some truly fresh unlabeled samples from $D$. The definitions above can naturally be extended to those settings, and we do so in the relevant sections.

\section{Evaluability for Amenable Distributions}
\label{sec:amenable_evaluable}
In this section, we will study the sample complexity of evaluation for amenable distributions.

\subsection{Amenable distributions are (weakly) evaluable}
We start with proving an upper bound on the sample complexity of evaluation for amenable distributions.

\begin{theorem}
\label{thm:smallsupp-main}
    For $k \in \mathbb{N}, \gamma \geq 0$, let $\dee_{k,\gamma}^\calX$ be the class of distributions that are $\{a_t\}$-amenable for some sequence with $a_k \leq \gamma$. Let $\cee_{all}^{\calY}$ be the class of all concepts mapping $\calX$ to $\Delta_\calY$. Let $s$ be any score function with range $[0,1]$. There is an evaluation algorithm that for any $\alpha > 0$, estimates the score  up to error $3\alpha+\gamma$ with $n = \frac{k\ln \frac k \delta}{\alpha} + \frac{2(k\calY\ln (1+\frac 1 \alpha) + \ln \frac 2 \delta ) }{\alpha^2}$ samples, except with probability $2\delta$.
\end{theorem}
\begin{proof}

    Let $D \in \dee_{k,\gamma}^\calX$. Recall that this implies that there is a set $U_k$ such that $p_D(U_k^c) \leq \gamma$, where $|U_k| \leq k$.

       The evaluator consists of several steps. It will be convenient to think of the sample from the distribution as a union of two separate samples $S_{n_1}$ and $S'_{n_2}$. The evaluator will use the first set of samples to compute the prediction on each $x_i$ in the first sample. It will then use the second sample to estimate the score on $D$, using a pessimistic default value for any new $x$'s in $S'_{n_2} \setminus S_{n_1}$. We give pseudo-code for the evaluator in \cref{alg:eval}.
    \begin{figure}[!h]
\begin{algorithm}[H]
\caption{Evaluator}\label{alg:eval}
    \begin{algorithmic}[1]
    \Require Unlabeled samples $S_{n_1} = \{x_1,\ldots,x_{n_1}\}$. Labeled samples $S'_{n_2} = \{(x'_1,y'_1),\ldots,(x'
    _{n_2}, y'_{n_2})\}$. Predictions $f(x)$ for each $x \in S_{n_1}$. Parameters $\alpha, \delta > 0$.
    \State Initialize $\hat{s}(x,y) = 1$ for all $(x,y) \in S'_{n_2}$.  \Comment{Start with default pessimistic value.}
    \State Initialize $T= \emptyset$.
    \For{$x \in S_{n_1}$}
    \State For each $y \in \calY$ s.t. $(x,y) \in S'_{n_2}$, set $\hat{s}(x,y) = s(f(x); y)$. \Comment{Update $\hat{s}$ for seen samples.}
    \State Update $T = T \cup \{x\}$.
    \EndFor
    \State Output estimate $\frac{1}{n_2} \sum_{(x,y) \in S'_{n_2}} \hat{s}(x,y)$.
    \end{algorithmic}
    \end{algorithm}
    \end{figure}

    In the algorithm, we denote by $T$ the set of $x$'s seen in the first sample $S_{n_1}$. We first prove that $T$ covers most of the mass of $D$ if $n_1$ is sufficiently large.
    \begin{lemma}\label{lem:T_covers}
        Suppose that $D \in \dee_{k,\gamma}^\calX$ and $n_1 \geq \frac{k\ln \frac k \delta}{\alpha}$. Then
        \begin{align*}
            \Pr_{S_{n_1} \sim D^{n_1}}[\sum_{x \in T} p_{D}(x) \geq 1 - \gamma - \alpha] \geq 1-\delta.
        \end{align*}
    \end{lemma}
    \begin{proof}
        Set $\tau = \alpha / k$, and define
        \begin{align*}
            T_{H} = \{x \in U_k : p_D(x) \geq \tau\} &;\;\; T_{L} = \calX \setminus T_H.%
        \end{align*}

        Then $p_{D}(T_L) \leq \sum_{x \in T_L\cap U_k} p_{D}(x) + p_D(U_k^c)  < k\tau +\gamma = \alpha+\gamma$. Moreover,
        \begin{align*}
            \Pr[T_H \not\subseteq T] &\leq \sum_{x \in T_H} \Pr[x \not\in T]\\
            &\leq \sum_{x \in T_H} (1-p_D(x))^{n_1}\\
            &\leq k\cdot (1-\tau)^{n_1}\\
            &\leq k\cdot e^{-\tau n_1}\\
            &\leq \delta
        \end{align*}
        when $n_1 \geq \tau^{-1}\ln \frac k \delta $. The claim follows.
    \end{proof}
    Next, we show that the score of any fixed $f$ is well-estimated on $S'_{n_2}$. The proof uses standard concentration inequalities  and is omitted.
    \begin{lemma}\label{lem:concentration}
        Let $\chi : \calX \rightarrow [0,1]$ and for a distribution $D$, let $\mu = \E_{x \sim D}[\chi(x)]$. Then
        \begin{align*}
            \Pr_{S_{n_2} \sim D^{n_2}}[|\frac{1}{n_2}\sum_{i=1}^{n_2} \chi(x'_i) - \mu| \geq \alpha] \leq 2e^{-\alpha^2 n_2 /2}
        \end{align*}
    \end{lemma}

    Let us fix an $x \in \calX$, and a $c \in \cee_{all}^\calY$. The expected score of a predictor $f$ when the input is $x$ is given by $\E_{y \sim c(x)}[s(f(x); y)] = \sum_{y \in \calY} c(x)_y \cdot s(f(x); y)$. Since the score of $f$ only depends (up to an additive $\gamma$) on the evaluation of $f$ on the $x$'s in the set $U_k$, the score of $f$ is fully determined by the $k\calY$ values $\{s(f(x); y) : x \in U_k, y \in \calY\}$.

    We will argue uniform convergence for the set of these values. Since $c(x) \in \Delta_{\calY}$, the function mapping the $k\calY$-dimensional score vector to the expected score is $1$-Lipschitz in $\ell_{\infty}$ norm. The straightforward net construction that rounds each $s(f(x); y)$ to a multiple of $\alpha$ yields a net of size $N_\alpha \eqdef \lceil \frac 1 \alpha \rceil ^{k\calY}$. Taking $n_2 \geq 2(\ln N_{\alpha} + \ln \frac 2 \delta)/\alpha^2$ then suffices to ensure uniform convergence over the net. By \cref{lem:T_covers}, the gap between $\hat{s}$ and $s$ leads to an additional error in expectation of at most $\alpha+\gamma$. The claim follows.

\end{proof}
For a sequence $\mathbf{a} = \{a_t\}_{t \in \mathbb{N}}$, let $\dee_{\mathbf{a}}$ be the family of $\{a_t\}_{t \in \mathbb{N}}$-amenable distributions. It follows that for any finite $\calY$, any bounded score function is $(\dee_{\mathbf{a}}, \cee_{all}^{\calY}, \mathbf{e})$-evaluable for $\mathbf{e}(n) = \min_{k} \Big(a_k + O\big(\frac{k\ln k}{n} + \sqrt{\frac{k\calY \ln (n/k\calY)}{n}}\big)\Big)$.

As we show in \cref{sec:tightness}, this bound is essentially tight in the worst case.  The next several theorems improve the bound in \cref{thm:smallsupp-main} under different assumptions.  The first improvement is for the case when the concept class consists only of deterministic maps. In this case the dependence on $|\calY|$ can be removed.
\begin{theorem}
\label{thm:det-concepts}
    For $k \in \mathbb{N}, \gamma \geq 0$, let $\dee_{k,\gamma}^\calX$ be the class of distributions that are $\{a_t\}$-amenable for some sequence with $a_k \leq \gamma$. Let $\cee_{det}^{\calY}$ be the class of all deterministic concepts mapping $\calX$ to $\calY$. Let $s$ be a score function with range $[0,1]$. There is an evaluation algorithm that for any $\alpha > 0$, estimates the score  up to error $3\alpha+\gamma$ with $n = \frac{k\ln \frac k \delta}{\alpha} + \frac{2(k\ln(1+\frac 1 \alpha)+ \ln \frac 2 \delta ) }{\alpha^2}$ samples, except with probability $2\delta$.
\end{theorem}
\begin{proof}
The proof is identical to that of \cref{thm:smallsupp-main}, except that we do a more refined analysis for the uniform convergence. For each $x$, we only care about $s(f(x), y)$ where $y=c(x)$, and the values $s(f(x), y')$ for other $y'$s are irrelevant for the uniform convergence argument. Thus we only need to build an $\alpha$-net for the $k$ relevant values. The claim follows.
\end{proof}

The next improvement is for the case when the score function is linear in the first argument, and allows us to reduce the dependence on $\calY$ from linear to logarithmic. This holds, e.g. for the common case of in-accuracy.

\begin{theorem}
\label{thm:smallsupp-linear}
    For $k \in \mathbb{N}, \gamma \geq 0$, let $\dee_{k,\gamma}^\calX$ be the class of distributions that are $\{a_t\}$-amenable for some sequence with $a_k \leq \gamma$. Let $\cee_{all}^{\calY}$ be the class of all concepts mapping $\calX$ to $\Delta_\calY$. Let $s$ be a score function with range $[0,1]$ that is linear in the first argument; i.e. $s(f(x); y) = \sum_j f(x)_j \cdot s(e_j; y)$. There is an evaluation algorithm that for any $\alpha > 0$, estimates the score  up to error $3\alpha+\gamma$ with $n = \frac{k\ln \frac k \delta}{\alpha} + \frac{2(k\ln \calY + \ln \frac 2 \delta ) }{\alpha^2}$ samples, except with probability $2\delta$.
\end{theorem}
\begin{proof}
The proof is identical to that of \cref{thm:smallsupp-main}, except that we do a more refined analysis for the uniform convergence. We do uniform convergence argument over {\em deterministic} predictions $f$, i.e. those where for each $x$, $f(x) = e_y$ for some $y\in \calY$ (depending on $x$). There are only $\calY^k$ such predictions on $U_k$ and thus $n_2 \geq \frac{2(k\ln \calY + \ln \frac 1 \delta ) }{\alpha^2}$ suffices to ensure uniform convergence on this set. Any prediction function $f$ is a convex combination of deterministic predictions. The linearity then allows us to accurately estimate the score of any prediction $f$.
\end{proof}

The next improvement holds for {\em sparse} score functions that for any prediction are zero except for a small number of labels (or symmetrically, are one except for a small number of labels). Our result is more general, and controls the sample complexity in terms of a bound on the $\ell_1$ norm of the score vector. By considering the score function $\hat{s}(f(x);y) = 1 - s(f(x); y)$, one can get a similar result for the case when $\sum_y (1-s(p; y)) \leq B$.

\begin{theorem}
\label{thm:smallsupp-sparse}
   For $k \in \mathbb{N}, \gamma \geq 0$, let $\dee_{k,\gamma}^\calX$ be the class of distributions that are $\{a_t\}$-amenable for some sequence with $a_k \leq \gamma$. Let $\cee_{all}^{\calY}$ be the class of all concepts mapping $\calX$ to $\Delta_\calY$. Let $s$ be a score function with range $[0,1]$ such that for all $p \in \Delta_{\calY}$, $\sum_y s(p; y) \leq B$. Then there is an evaluation algorithm that for any $\alpha > 0$, estimates the score  up to error $4\alpha+\gamma$ with $n = \frac{k\ln \frac k \delta}{\alpha} + \frac{2 kB\ln ((1+\alpha)e\calY/B)}{\alpha^3} + \frac{2\ln \frac 2 \delta }{\alpha^2}$ samples, except with probability $2\delta$.
\end{theorem}
\begin{proof}
The proof is identical to that of \cref{thm:smallsupp-main}, except that we do a more refined analysis for  uniform convergence. Recall that it suffices to build a ``net'' for the set of vectors $\{s(f(x); y) : x \in U_k, y \in \calY\}$. Let us first fix the concept $c$. For a fixed $x$, and for any $y$ such that $c(x)_y \leq \frac \alpha B$, we can round $s(f(x);y)$ to zero; the total error introduced due to this is at most
\begin{align*}
    \sum_y c(x)_y \mathbf{1}(c(x)_y \leq \frac \alpha B) s(f(x);y) &\leq     \sum_y \frac \alpha B s(f(x);y)\\ &\leq
      \frac \alpha B \sum_y  s(f(x);y)\\
      &\leq \alpha.
\end{align*}
Thus we only need to consider in our net the points $y$ for which $c(x)_y > \frac{\alpha}{B}$, and there are at most $\frac B \alpha$ such $y$. Thus it suffices to include in our net all vectors where the number of non-zero values amongst $s(x,y)$ for a given $x$ is at most $B/\alpha$.
This allows us to improve the bound on $|N_\alpha|$ to ${\calY \choose {B}/{\alpha}}^k \cdot \ceil{\frac 1 \alpha}^{\tfrac {kB} \alpha} \leq ((1+\alpha)e\calY/B)^{\tfrac {kB} \alpha}$.
\end{proof}
Our final result here improves the bounds when the score is small.
\begin{theorem}
\label{thm:small-supp-zero-error}
Suppose that the predictor $f$ has expected score at most $\mu$ on the distribution $D$. Then for any $\alpha \leq \mu$:
\begin{itemize}
\item In the setting of \cref{thm:smallsupp-main}, the accuracy bound holds as long as $n \geq \frac{k\ln \frac k \delta}{\alpha} + \frac{3\mu(k\calY\ln (1+\frac 1 \alpha) + \ln \frac 2 \delta ) }{\alpha^2}$.
\item In the setting of \cref{thm:smallsupp-linear}, the accuracy bound holds as long as $n \geq \frac{k\ln \frac k \delta}{\alpha} + \frac{3\mu(k\ln \calY + \ln \frac 2 \delta ) }{\alpha^2}$.
\item In the setting of \cref{thm:smallsupp-sparse}, the accuracy bound holds as long as $n \geq \frac{k\ln \frac k \delta}{\alpha} + \frac{3\mu kB\ln ((1+\alpha)e\calY/B)}{\alpha^3} + \frac{3\mu\ln \frac 2 \delta }{\alpha^2}$.
\end{itemize}
\end{theorem}
\begin{proof}
    We use the multiplicative form of the Chernoff bound instead of the additive error form. Up to constants, it suffices to use the following standard version:
    \begin{lemma}
    \label{lem:concentration-mult}
        Let $\chi : \calX \rightarrow [0,1]$, and for a distribution $D$, let $\mu = \E_{x \sim D}[\chi(x)]$. Then for $\alpha \leq \mu$
        \begin{align*}
            \Pr_{S_{n_2} \sim D^{n_2}}[|\frac{1}{n_2}\sum_{i=1}^{n_2} \chi(x'_i) -  \mu| > \alpha] \leq 2e^{-\alpha^2 n_2/3\mu}
        \end{align*}    \end{lemma}
        Plugging this version of Chernoff bounds in lieu of \cref{lem:concentration} yields the claimed bounds.
\end{proof}

\begin{remark}
    These results can be naturally translated to bounds on $\mu$ given a certain estimate output by \cref{alg:eval}. Thus for example, let's consider the case of $0$-$1$ accuracy for a binary prediction task for distributions in $\dee_{k,0}^\calX$, i.e. distributions with support size $k$. Then \cref{thm:smallsupp-linear} tells us that the output of the algorithm is accurate up to an additive $O(\frac{k \ln k/\delta}{n} + \sqrt{\frac{k+\ln \tfrac 2 \delta}{n}})$. If the estimate happens to be zero, then \cref{thm:small-supp-zero-error} allows us to improve this bound to $O(\frac{k \ln k/\delta}{n} + \frac{k+\ln \tfrac 2 \delta}{n})$.
\end{remark}

These bounds with $\frac k {\alpha^2}$ type dependence are significantly worse than the $\frac 1 {\alpha^2}$ dependence one would need when fresh (labeled) test samples are available. In our proof, this dependence comes from the fact that we build on uniform convergence for the family of all functions. In \cref{sec:tightness}, we show that this dependence is unavoidable. Not surprisingly, the lower bound is proven by constructing $f$ that not only ``memorize'' the samples that both Lauren and Evelyn see (which would lead to a $k$ dependence), but also carefully makes errors on examples whose frequencies look similar in the sample, but differ in the distribution.

\subsection{Amenable distributions are (strongly) evaluable}
In the last section, we argued that distributions that are close to being small support are weakly evaluable: if the underlying distribution $D$ indeed is amenable, then the evaluation algorithm provides an accurate estimate. We next argue that we can strengthen this to strong evaluability. To do so, we need to produce an estimate that is accurate when the input distribution is amenable, but is a valid upper bound even when the input distribution may not be amenable.

We can do this, intuitively, by estimating the amenability parameters of the distribution from samples. We define a subroutine that for $n$ samples from an arbitrary distribution $D$, computes estimates $\abar_k$ that approximate the amenability parameters $a_k$ of the distribution, up to an additive error that grows as $\sqrt{\frac{k}{n}}$, up to logarithmic factors. Given these estimates, the algorithm essentially computes the bounds one would get for each $k$ and selects the tightest of those bounds. As each of these (up to $n$) bounds is valid, the algorithm gives always valid bounds. When $D \in \dee_{k,\gamma}^{\calX}$, the bounds from the estimates of $a_k$ are sufficiently close.

\begin{theorem}
\label{thm:smallsupp-main-strong}
   For $k \in \mathbb{N}, \gamma \geq 0$, let $\dee_{k,\gamma}^\calX$ be the class of distributions that are $\{a_t\}$-amenable for some sequence with $a_k \leq \gamma$.
   Let $\cee_{all}^{\calY}$ be the class of all concepts mapping $\calX$ to $\Delta_\calY$. Let $s$ be any score function with range $[0,1]$. Then there is an evaluation algorithm that for any distribution $D$, provides upper and lower bound estimates for the expected score on $D$ that are valid except with probability $\delta$. If the distribution $D \in \dee_{k,\gamma}^{\calX}$, the upper and lower bounds differ by at most $\gamma + O\big(\frac{k\ln \frac n \delta}{n} + \sqrt{\frac{k\calY\ln n + k\ln \frac n \delta}{n}}\big)$.
\end{theorem}
\begin{proof}

       The evaluator is similar to \cref{alg:eval}, except that instead of taking $k$ and $\gamma$ as inputs, it estimates the parameters $a_k$ of the distribution for all $k$. It then uses the results of the previous section to construct for each $k$, upper and lower bounds on the expected score on $D$. It outputs the tightest of these bounds.

       We start with an amenability estimation algorithm. Given $n$ samples, this algorithm constructs for each $k$, a reasonable candidate $P_k$ for the set $U_k$. It then uses $n$ fresh samples to estimate the mass the distribution assigns to $P_k^c$. We argue that these estimates are reasonably accurate.
           \begin{figure}[!h]
\begin{algorithm}[H]
\caption{Amenability Estimator}\label{alg:amen_est}
    \begin{algorithmic}[1]
    \Require Samples $S_{n} = \{x_1,\ldots,x
    _{n}\}$, $S'_{n} = \{x'_1,\ldots,x'
    _{n}\}$, $S''_{n} = \{x''_1,\ldots,x''
    _{n}\}$. Parameter $\delta > 0$.

    \State Let $T$ be the support of $S_n$.
    \For {$x \in T$} \State Estimate $p'(x) = \frac 1 n \sum_i \one(x'_i = x)$ using $S'_n$.\label{alg:estimate}
    \EndFor
    \State Sort $T$ in decreasing order of $p'(\cdot)$.\label{alg:sort}
    \For {$k = 1,\ldots, |T|$}
    \State Let $P_k$ denote the first $k$ elements of $T$
    \State Compute estimate $q_k = \frac 1 n \sum_{i} \one(x''_i \not\in P_k)$.\label{alg:estimatefin}
    \EndFor
    \State Output estimates $q_1,\ldots q_{|T|}$.
    \end{algorithmic}
    \end{algorithm}
    \end{figure}
The following lemma follows from standard concentration inequalities.
\begin{lemma}
    Let $p'(x)$ be the estimate of $p(x)$ computed in \cref{alg:estimate} of the Amenability Estimator above. Then for any $x \in T \cup U_n$, $|p'(x) - p(x)| \leq C(\sqrt{\frac{p(x) \ln \frac n \delta}{n}} + \frac{\ln \frac n \delta}{n})$ for $C = \sqrt{6}$, except with probability $\delta^2/n^2$.
\end{lemma}
Without loss of generality, let $U_k$ be the set of the $k$ heaviest elements under $D$, so that $U_1 \subseteq U_2 \subseteq \cdots \subseteq U_n$. Suppose that this event in the conclusion of the above lemma happens for all $x \in T \cup U_n$. Let $P_k$ be the set of the first $k$ $x$'s after the sorting step in \cref{alg:sort}. Then it follows that
\begin{align*}
    \sum_{x \in P_k} p(x) &\geq  \sum_{x \in P_k} (p'(x) - C\sqrt{\frac{p(x) \ln \frac n \delta}{n}} - C\frac{\ln \frac n \delta}{n})\\
    &\geq \sum_{x \in U_k \cap T} p'(x) - \sum_{x\in P_k} C\sqrt{\frac{p(x) \ln \frac n \delta}{n}} - \sum_{x\in P_k} C\frac{\ln \frac n \delta}{n}\\
    &\geq \sum_{x \in U_k \cap T} p(x) - \sum_{x\in U_k} C\sqrt{\frac{p(x) \ln \frac n \delta}{n}} - \sum_{x\in P_k} C\sqrt{\frac{p(x) \ln \frac n \delta}{n}} - \sum_{x\in U_k} C\frac{\ln \frac n \delta}{n} - \sum_{x\in P_k} C\frac{\ln \frac n \delta}{n}\\
    &= \sum_{x \in U_k \cap T} p(x) - C\sqrt{\frac{\ln \frac n \delta}{n}} (\sum_{x\in U_k} \sqrt{p(x)} +  \sum_{x \in P_k} \sqrt{p(x)}) - 2Ck\frac{\ln \frac n \delta}{n}\\
    &\geq \sum_{x \in U_k \cap T} p(x) - C\sqrt{\frac{4k\ln \frac n \delta}{n}} - 2Ck\frac{\ln \frac n \delta}{n}
\end{align*}
Here we have used Cauchy-Schwarz in the last step. Moreover, applying the argument of \cref{lem:T_covers} to $S_n$ shows that, except with probability $\delta$, $\sum_{x \in U_k \setminus T} p(x) \leq \frac{2k\ln \frac n \delta}{n}$ for all $k$ simultaneously, so that $\sum_{x \in P_k} p(x) \geq \sum_{x \in U_k} p(x) - C\sqrt{\frac{4k\ln \frac n \delta}{n}} - (2C+2)k\frac{\ln \frac n \delta}{n}$. It follows that the quantity being estimated in \cref{alg:estimatefin}, namely the mass outside $P_k$, is at most $a_k$ plus an additive $O(\sqrt{\frac{4k\ln \frac n \delta}{n}})$ whenever $n > 4k\ln \frac n \delta$. Since $|P_k| = k$, this mass is also at least the smallest mass outside any set of size $k$. By concentration, $q_k$ estimates it up to an additive $O(\sqrt{\frac{\ln \frac n \delta}{n}})$. Thus $q_k + O(\sqrt{\frac{\ln \frac n \delta}{n}})$ is a valid bound on the mass of $D$ outside some set of size $k$, and exceeds $a_k$ by at most an additive $O(\sqrt{\frac{4k\ln \frac n \delta}{n}})$. Since the failure probability is small enough, we can do a union bound over all $k$. Since we use the best of the upper and lower bounds, their difference is smaller than the smallest of the differences, and hence the claim follows.

 \end{proof}

We thus get strong evaluability using the evaluator that outputs the upper bound, with $\err(n)$ growing as $\gamma + O\big(\frac{k\ln \frac n \delta}{n} + \sqrt{\frac{k\calY\ln n + k\ln \frac n \delta}{n}}\big)$. We remark that our positive results hold for the class of all concepts, whereas the negative results we show next hold essentially for any $\cee$.

\subsection{Evaluability requires amenability}
We next argue that amenability is needed for evaluability in the black-box sense of \cref{def:evaluability}.

\begin{theorem}
\label{thm:eval_needs_amenability}
Let $D$ be a distribution on $\calX$ and $c$ a concept from $\calX \rightarrow \calY = \{0,1\}$. For $\tau > 0$, let $s$ be a score function such that for each $y \in \calY$, there exist $g_y, b_y \in \Delta_\calY$ with $s(b_y; y) - s(g_y; y) \geq \tau$. Suppose that for a function $\err : \mathbb{N} \rightarrow [0,1]$, $s$ is $(\{D\}, \{c\}, \err)$-evaluable. Then $D$ is $\{2\tau^{-1}\err(k)\}_{k \in \mathbb{N}}$-amenable.
\end{theorem}
\begin{proof}
Given $k$, we will construct a set $U_k$ and two prediction functions $f^g$ and $f^b$, that look indistinguishable to the evaluation algorithm. Our construction will ensure that the gap between the expected scores of $f^g$ and $f^b$ is at least $\tau \cdot \Pr_{x \sim D}[x \not\in U_k]$. Since the evaluator ensures error at most $\err(k)$ on every $f$, its output is within $\err(k)$ of the scores of $f^g$ and $f^b$. This will imply the claim. We now give details.

We give a randomized construction. Let $S_k$ be a sample of size $k$ from $D$, and let $U$ be the support of $S_k$. Clearly $|U| \leq k$ with probability $1$. We define the following (deterministic) prediction functions:
\begin{align*}
f^g_U =
\left\{ \begin{array}{ll}
e_{c(x)} & \mbox{ if } x \in U\\
g_{c(x)} & \mbox{ otherwise}\\
\end{array}\right.
\;\;\;\;;&\;\;\;\;
f^b_U =
\left\{ \begin{array}{ll}
e_{c(x)} & \mbox{ if } x \in U\\
b_{c(x)} & \mbox{ otherwise}\\
\end{array}\right..
\end{align*}
It is immediate that since $f^g_U$ and $f^b_U$ are identical on $U$, the evaluator cannot distinguish between these two prediction functions. Let $\bar{s}_U$ be the estimate produced by the evaluation algorithm. By the error bound of the evaluation algorithm, we have
\begin{align*}
    |\E_U[\bar{s}_U] - \E_U[\E_{x \sim D}[s(f^g_U(x),c(x))]]| \leq \mathbf{e}(k);\\
    |\E_U[\bar{s}_U] - \E_U[\E_{x \sim D}[s(f^b_U(x),c(x))]]| \leq \mathbf{e}(k).
\end{align*}
Thus by triangle inequality, we have
\begin{align*}
    | \E_U[\E_{x \sim D}[s(f^b_U(x),c(x))]] - \E_U[\E_{x \sim D}[s(f^g_U(x),c(x))]]| \leq 2\mathbf{e}(k).
\end{align*}
On the other hand, by construction, for any $U$, we have
\begin{align*}
    \E_{x \sim D}[s(f^b_U(x),c(x))] - \E_{x \sim D}[s(f^g_U(x),c(x))] &\geq \tau \cdot \Pr_{x \sim D}[x \not\in U].
\end{align*}
It follows that
\begin{align*}
    \E_U[\Pr_{x \sim D}[x \not\in U]]
    &\leq \tau^{-1} \cdot \E_U[\E_{x \sim D}[s(f^b_U(x),c(x))] - \E_{x \sim D}[s(f^g_U(x),c(x))]]\\
    & \leq 2\tau^{-1} \mathbf{e}(k).
\end{align*}
Thus there exists a $U$ of size at most $k$ such that $\Pr_{x \sim D}[x \not\in U] \leq  2\tau^{-1} \mathbf{e}(k)$. The claim follows.
\end{proof}

While amenability may seem like a strong restriction, our results show that amenability is an exact characterization.  The lower bound above says that modulo trivialities, {\bf no} score function is evaluable on {\bf any} non-amenable distribution. Assumptions on score functions and/or other assumptions on the distribution will not help get around this impossibility. Thus if $D$ is amenable, then we can do strong evaluation for any concept and for any score function. If $D$ is not amenable, then there is no non-trivial concept class and no non-trivial score function for which even weak evaluation is possible. In other words, amenability of $D$ is the whole game.

\section{Evaluability with Additional Access}
\label{sec:oracle_access}

In this section, we empower Evelyn to query the prediction model $f$ on additional inputs not in the sample $S_n$. We first make some simple observations. Recall from \cref{thm:estimation} that $O(\frac{\ln \frac 1 \delta}{\alpha^2})$ fresh labeled samples suffice to evaluate for any score function and any distribution. It follows that if Evelyn has $O(\frac{\ln \frac 1 \delta}{\alpha^2})$ fresh {\em labeled} samples, she can evaluate to accuracy $\alpha$.

However labeled examples may be difficult to procure, and in the rest of this section, we will ask what one can do with some additional fresh {\em unlabeled} samples from $D$. We start with formally defining this setting.

We consider a small modification of the game defined in \cref{sec:defn}. We have a distribution family $\dee$, a concept class $\cee$ and a score function $s$. Nature selects a $D \in \dee$ and a $c \in \cee$ such that $D$ is consistent with $c$. A sample of $n$ examples $S_n = \{(x_1,y_1),(x_2,y_2),\ldots,(x_n,y_n)\}$ is drawn i.i.d. from $D$ and provided to \Lauren\ and \Evelyn. In addition, an independent sample of $m$ unlabeled examples $T_m = \{x'_1, x'_2,\ldots, x'_m\}$ is provided to \Evelyn. \Lauren\ uses the $n$ samples, and potentially any other information she may have, to generate a predictor $f$. \Evelyn\ gets access to $S_n$, $T_m$ and to the evaluation of $f$ on each $x_i \in S_n$ and on any additional set $T'_m$ of $m$ points\footnote{In this definition, we have set the number of unlabeled samples Evelyn receives to be equal to the number of additional queries for simplicity. These are however two different resources and a more general definition should treat them separately.} she may choose.
She must produce an estimate $E(S_n, f(S_n), T_m, f(T'_m))$ of the score $s(f; D)$.

\begin{definition} [Evaluability with Additional Unlabeled Access]
  Let \dee\ be a class of distributions, and \cee\ a concept class. Let $\err: \mathbb{N} \times \mathbb{N} \rightarrow [0,1]$. We say a score function $s(\cdot, \cdot)$ is $(\dee, \cee, \err)$-evaluable with additional access if there is an algorithm $E$ such that:
  \begin{itemize}
  \item {\bf Completeness}: For every $c \in \cee$ and every $D \in \dee$ consistent with $c$, it is the case that
  \begin{align*}
      \E_{S_n \sim D^n}[\sup_f \left( \E_{T_m \sim D^m} [E(S_n,f(S_n), T_m, f(T'_m))]- s(f;D)\right)] &\leq   \err(n,m).
  \end{align*}
      \item {\bf Soundness}: For every $c \in\cee$ and every $D \in \dee$ consistent with $c$, it is the case that
  \begin{align*}
      \E_{S_n \sim D^n}[\inf_f \left( \E_{T_m \sim D^m}[E(S_n,f(S_n), T_m, f(T'_m))] - s(f;D)\right)] &\geq  - \err(n,m).
  \end{align*}
  \end{itemize}
  If $s$ is $(\dee,\cee,\err)$-evaluable with additional access for some $\err$ with $\lim_{n\rightarrow \infty}\err(n,n)=0$, we say that it is $(\dee,\cee)$-evaluable with additional access. In this case, we say $s$ is evaluable with $n(\alpha)$ labeled samples and $m(\alpha)$ fresh unlabeled samples if $n(\cdot)$ and $m(\cdot)$ satisfy $\err(n(\alpha), m(\alpha)) \leq \alpha$  for all $\alpha>0$.
\end{definition}

We first observe that irrespective of the number of unlabeled points it may use, an evaluator for inaccuracy must use at least $VCDim(\cee)$ labeled points.
\begin{theorem}
\label{thm:vcdimneeded}
    Suppose that for $\calY = \{0,1\}$, there is an algorithm for estimating inaccuracy over $(D,\cee)$ which makes $L$ labeled and $U$ unlabeled queries. If $L$ is such that with constant probability over a sample $S= \{x_1,\ldots,x_L\}$ drawn i.i.d. from $D$, the value of $c$ on the support of $D$ is not fully determined by $c(x): x \in S$, then the evaluator's error cannot converge to $0$ (as $U$ increases).
\end{theorem}
\begin{proof}
Without loss of generality, suppose that $\cee$ is finite (or else replace $\cee$ by a suitably fine net). Consider the setting where $c \in \cee$ is chosen uniformly at random. By the assumption on $L$, with constant probability over the sample $S$,  conditioning on the values $\{c(x) : x \in S\}$ does not fix $c$. Thus there is an $x \in S^c$ with non-zero probability mass and non-zero entropy in $c(x)$. Consider the learner that picks $f$ to be a random $c$ that is consistent with the labels on $S$. When $c=f$, the inaccuracy is zero, and it is non-zero otherwise. Furthermore, from the point of view of the evaluator, any two $c$'s consistent with the labels on $S$ are equivalent: the learner behaves identically on the two, and hence all queries made on unlabeled points will yield the same result. It follows that the evaluator's error cannot converge to zero.
\end{proof}

Generically, one would need to fix $c(x)$ on at least $VCDim(\cee)$ points to determine $c$, so unless $D$ and $\cee$ are aligned in an adversarial way, the above lower bound is $\Omega(VCDim(\cee))$.
As an example, if $D$ is uniform on $\calX=\{0,1\}^d$ and $\cee$ is the class of parity functions on $\calX$, then one needs $\Omega(d)$ labeled samples to determine a parity, and hence any evaluator needs $\Omega(d)$ labeled samples. Thus we will allow access to $O(VCDim(\cee))$  labeled points, and ask when the evaluator can get away with a small number of additional unlabeled points. A natural approach is to compare the model $f$ to the data on the labeled points, and then {\em test} that $f$ is (close to) a function in $\cee$. Indeed results in property testing allow us to efficiently determine if $f$ is close to a function in $\cee$ (using the fresh unlabeled samples). In the extreme case, if $f$ agrees with $c$ on the $L$ labeled samples, and $f$ is in $\cee$, uniform convergence for $\cee$ implies that $f$ must have low error. One may conjecture that  uniform convergence and proximity to $\cee$  is sufficient to ensure low error. Unfortunately, this approach fails. Consider an $f$ that agrees with $c$ on the labeled set of points, but is equal to a different $c' \in \cee$ on every other point. This $f$ will have zero error on the labeled points, will be really close to $\cee$, and yet have large error on $D$. Testing $f$ for closeness to $\cee$ is thus not sufficient for accurate evaluation.

We instead build on a richer object called a {\em self-corrector}. A self corrector for a concept class $\cee$ and a distribution $D$ allows one to compute, at a given point $x$, a function $g(x)$ such that $g$ is close to being in $\cee$, and is close to $f$. While the function $c_f \in \cee$ that is closest to $f$ satisfies this property, a local self-corrector allows for computation of $g(x)$ at a point $x$ while making a small number of queries to $f$. We prove the following general result that follows this recipe.

\begin{theorem}\label{thm:corrector_to_eval}
    Suppose that for class $\dee$ of distributions over $\calX$ and a class $\cee$ of functions from $\calX$ to $\{0,1\}$, we have a {\em self-corrector}  with the following properties:
    \begin{itemize}
        \item Given $x\in \calX$, sample access to $D\in \dee$, and oracle access to $f$, the self corrector draws $n_{SC}$ samples from $D$, queries $f$ at $n_{SC}$ points in $\calX$, each distributed according to $D$, and w.p. $(1-\delta)$ outputs a value $g(x)$.
        \item The function $g : \calX \rightarrow \{0,1\}$ lies in $\cee$.
        \item The distance $d_D(f, g)$ is bounded by $\kappa \cdot \inf_{c \in \cee} d_D(f, c)$ except with probability $\delta$.
    \end{itemize}
    Then for any $\alpha>0$, and any bounded score function $s$, there is an evaluator that gets $L = \tilde{O}(VCDim(\cee)/\alpha^2)$ labeled samples from $D \in \dee$ (shared with the learner, and labeled using a $c \in \cee$) and $L \cdot n_{SC} + O(\frac{\ln 1/\delta}{\alpha^2})$ fresh unlabeled samples from $D$. It queries $f$ on at most $Ln_{SC}$ points in $\calX$ and outputs an estimate $\hat{s}$ that w.p. $(1-\delta)$ satisfies:
     \begin{align*}
         s(f; D) \leq \hat{s} \leq s(f;D) + O(\alpha + \kappa d_D(f, \cee)).
     \end{align*}
\end{theorem}
\begin{proof}
We start with a version of the standard uniform convergence argument for VC classes. Suppose that $L$ is of size $L \geq C\cdot\frac{VCDim(\cee)\log \frac 1 \alpha + \log \frac 1 \delta}{\alpha^2}$, and $s$ is a bounded score function.  Then for an appropriate constant $C$, it holds that for all $c \in \cee$, $|s(c; D) - s(c; L)| \leq \alpha$. We refer the reader to a standard textbook on learning theory (e.g.~\citep{Shalev-Shwartz_Ben-David_2014}) for a proof of this claim. We now write:
\begin{align*}
    s(f; D) &= \E_{(x,y) \sim D}[s(f(x); (x,y))]\\
    &\leq \E_{(x,y) \sim D}[s(g(x); (x,y))] + d_D(f, g)\\
    &\leq (\E_{(x,y) \in L}[s(g(x); (x,y))] + \alpha) + d_D(f, g).
\end{align*}
Similarly, we can write:
\begin{align*}
    s(f; D) &= \E_{(x,y) \sim D}[s(f(x); (x,y))]\\
    &\geq \E_{(x,y) \sim D}[s(g(x); (x,y))] - d_D(f, g)\\
    &\geq (\E_{(x,y) \in L}[s(g(x); (x,y))] - \alpha) - d_D(f, g).
\end{align*}
The first term $\E_{(x,y) \in L}[s(g(x); (x,y))]$ can be computed using the self-corrector. The last term $d_D(f, g)$ can also be estimated by evaluating $f$ and $g$ on a fresh set of $O(\frac{\ln\frac{1}{\delta}}{\alpha^2})$ samples from $D$. Thus these upper and lower bounds can be computed up to accuracy $O(\alpha)$. This implies the claim.

We also observe that for scores such as inaccuracy, we can upper bound $d_D(f,\cee)$ by $s(f; D)$. In this case, we get a multiplicative-additive approximation to the score.
\end{proof}

We remark that in typical applications, one is interested in models $f$ that have low inaccuracy. This implies that the distance $d_D(f, \cee)$ is bounded by the inaccuracy. This in turn implies that our approximation error for any other score is small. As self-correctors typically need assumptions on the distribution family, this only gives us a weak evaluator. If in addition, we had a tester for the distribution family, we could get a strong evaluator in a straightforward way. However, distribution testing over large domains typically needs many samples and thus the sample complexity benefits of the additional access for strong evaluability are much more limited. Thus we restrict ourselves to weak evaluability in this section.

We state some simple corollaries of this result, showing settings where evaluation can be done with very few fresh samples. The first result builds on the line of work on linearity testing and self-correctors for linearity~\citep{BlumLR93}. We use here a result from \citet{KoppartyS09}.
\begin{corollary}
    Let $\dee$ be the family of product distributions on $\{0,1\}^d$ satisfying $\Pr_{x \sim D}[x_i = 1] \in [\frac 1 {10}, \frac 9 {10}]$ for each $i \in [d]$. Let $\cee$ be the concept class of parity functions from $\{0,1\}^d$ to $\{0,1\}$. Let $s$ be a bounded score function. Given $\tilde{O}(\frac d {\alpha^2})$ labeled samples from $D$, and $\tilde{O}(\frac d {\alpha^2})$ unlabeled samples from $D$, there is an efficient evaluator that for any $f$, computes an estimate $\hat{s}$ satisfying $s(f; D)  \leq \hat{s} \leq O(s(f;D) + \alpha + d_D(f, \cee))$.
\end{corollary}
\begin{proof}
The proof follows from the self-corrector from~\citep{KoppartyS09}. One can verify that generating query points for the self corrector can be done using only sample access to $D$.
\end{proof}
The above corollary continues to hold if an $\alpha$ fraction of the labels are incorrect. Note that in this case, learning $c$ is believed to be computationally hard, but evaluation is efficient. The result holds for the larger class of {\em uniformly correlatable} distributions, that includes symmetric distributions and low-bias distributions. We refer the reader to \citet{KoppartyS09} for additional details.

A similar result holds for $\cee$ being the class of $k$-juntas, and more generally low-degree polynomials over $\mathbb{F}_2$, and $\dee$ being the set of distributions close to the uniform distribution. This builds on self-correctors for low degree polynomials and juntas~\citep{AlonKKLR03, AlonW12}.
\begin{corollary}
    Let $\dee$ be the family of distributions on $\{0,1\}^d$ satisfying $p_D(x) \leq 2\cdot2^{-d}$. Let $\cee$ be the concept class of $k$-juntas from $\{0,1\}^d$ to $\{0,1\}$. Let $s$ be a bounded score function. Given $\tilde{O}(\frac {2^k + k \log d} {\alpha^2})$ labeled samples from $D$, and $O_k(\poly(\log d/\alpha))$ unlabeled samples from $D$, there is an efficient evaluator that for any $f$, computes an estimate $\hat{s}$ satisfying $s(f; D) \leq \hat{s} \leq O(s(f;D) + \alpha + d_D(f, \cee))$.
\end{corollary}
\begin{proof}
$k$-juntas have VC dimension $O(2^k + k\log d)$, and being degree at most $k$ polynomials, they can be self-corrected using constant (as a function of $k$ and $\alpha$) evaluations for the uniform distribution. Our assumption on $\dee$ implies that distance under $D$ is upper bounded by a constant times the distance under the uniform distribution.
\end{proof}

This result can be extended to degree-$k$ polynomials with a bounded number of monomials, as that is sufficient for the VC dimension bound.

Finally, we show that evaluation for certain scores can be significantly harder than ``agnostic learning'' for those scores. Indeed there are score functions that are easy to approximately minimize in a trivial way. For example,
the inaccuracy under abstentions score of \cref{exa:abstain} can be trivially minimized by a classifier that always abstains and thus has sample complexity $0$. However, given an arbitrary $f$, evaluating its score requires a non-zero number of samples. Thus evaluating a score function can be considerably harder than minimizing it.

We remark that for inaccuracy, evaluation with VCDim many labeled samples may seem futile: if the goal is to build a good classifier, the evaluator might as well just learn a new classifier.  There are several reasons why evaluation may still be interesting. Firstly, often the model of interest (such as a foundation model) is a common model used for several separate tasks, and deploying a separate new model for each task may not be feasible. Secondly, the cost of training a new model may be large. This is often true in practice, and there are several classes of interest where agnostic learning is computationally hard. Thirdly, this is not an option when we want to evaluate the same model along several axes, i.e. want to evaluate several score functions. This is commonly the case in real-world deployments, making the train afresh approach inadmissible.

\section{Conclusions}\label{sec:conclusions}
In this work, we have defined a notion of evaluability, when the assumptions of a fresh test set may not hold. Our notion is general, and encompasses a large set of score functions that may be important to determine fitness for use for a model. We have studied this problem in a few different models. We show that in the most basic model, when the evaluator only looks at the model output on a public benchmark, accurate evaluation is possible if and only if the underlying distribution of interest is amenable (modulo trivialities). We then show that allowing the ability to query the model on additional points can make the evaluator more powerful in certain settings, via a surprising connection to self-correctors studied in program testing.

Several natural questions remain. Our main negative result~\cref{thm:eval_needs_amenability} builds on a very adversarial learner, and it would be natural to ask if there are reasonable restrictions one can place on the learner that one can hope to verify. More broadly, our evaluation algorithms need no cooperation from the model, and we leave open the question of understanding the power of evaluators when the model may be designed to be better evaluated, and/or can be part of the process of evaluation. As a simple example, a model that proves mathematical theorems can help by providing a formalization in a verifiable language such as LEAN. More generally, an evaluator can engage in an interactive protocol with the model or model generator~\citep{AmitGPR24}.

Our work studies the class of score functions that are expectations of per-example scores. This includes a large class of interesting scores, and allows estimation of others. There may however be other scores of interest that are estimable from samples but not quite from expectations, and we leave the study of those to future work. This restriction also ignores additional factors that may determine fitness for use. E.g. our framing assumes a fixed distribution, and thus avoids questions of robustness to distribution shift. Distribution shift may occur due to natural shift in the distribution,  rational best-response to the use of the model, or adversarial attacks meant to exploit the weaknesses of the model. Our work also ignores the question of the interpretation of an evaluation metric. If success on a certain metric is interpreted as hard-to-quantify notions such as "ability to reason", such interpretations are always subject to the validity of the task as measuring those. These questions of {\em construct validity}~\citep{CronbachM55} are beyond the scope of this work.

We saw that self-correction is a powerful tool that allows evaluation of models even without access to any fresh labeled data. This motivates the question of designing better self-correctors for a larger set of concept classes. While a lot of work on self-correctors studies function classes over discrete alphabets, some recent work has studied self-correctors over reals, for some classes of distributions~\citep{FlemingY20, AroraBFKY23, AroraKM25, AroraBJS26}. We expect some of these results to be translatable to evaluability results. Our framework also allows some wiggle room here: an ``improper'' self-corrector where $g$ can lie in a slightly larger function class may suffice for this application. Further, it is unclear if self-correction is necessary, and there may be other approaches to evaluability.

\section*{Acknowledgements}
Many thanks to Parikshit Gopalan for useful discussions, and for a pointer to work on tolerant linearity testing. Thanks also to Adam Smith for useful discussions.

\bibliographystyle{plainnat}
\bibliography{refs}

\appendix
\section{Tightness of Sample Complexity Bounds}
\label{sec:tightness}
The proof of the tightness of the bounds follows, morally, from the lower bound for discrete distribution estimation. Technically however, we seem to need to unroll that proof and work from first principles. We will use the following score function:
\begin{align*}
    s(p; y) = \left\{\begin{array}{ll} 0 & \mbox{ if $p_y \geq \frac 1 {|\calY|}$} \\ 1 & \mbox{ otherwise} \end{array}\right.
\end{align*}

Suppose that for a domain $[d]$ where $d=k\calY$, we draw $P$ from a prior $Dirichlet(\beta, \beta, \ldots, \beta)$. Then given a histogram $n_1,\ldots, n_d$ sampled from $P$, the posterior on $P$ is $Dirichlet(\beta+n_1, \beta+n_2,\ldots, \beta+n_d)$.

We first define subsets $H$, $L$ of $[d]$ based on $P$ and the histogram:
\begin{align*}
    H &= \cup_{n \in \mathbb{N}} H_n\\
    \mbox{where } H_n &= \{j : n_j = n \mbox{ and } P_j \geq Median(Dirichlet(n_1+\beta,\ldots,n_d+\beta)_j)\}.\\
    L &= \cup_{n \in \mathbb{N}} L_n\\
    \mbox{where } L_n &= \{j : n_j = n \mbox{ and } P_j \leq Median(Dirichlet(n_1+\beta,\ldots,n_d+\beta)_j)\}.
\end{align*}
Informally, $H$ corresponds to the coordinates where $P$ is larger than the median, where we take the median conditioned on each observed count. Symmetry implies that the distributions of $H_n$ and $L_n$ are identical, and thus the distributions of $((n_1,\ldots,n_d), H)$ and $((n_1,\ldots,n_d), L)$ are identical. The marginals of $P_j$ are a Beta distribution. The large mean absolute deviation of a Beta distribution, coupled with closeness of mean and median of the Beta distribution, imply that (in expectation over $P$ drawn from the posterior), the difference in $P(H)$ and $P(L)$ is large. Indeed it follows from the work of \citet[Appendix D]{pmlr-v40-Kamath15} that for $\beta \geq 1$, and large enough $n$, there is a constant $C$ such that in expectation,
\begin{align*}
    \sum_{j \in H} P_j - \sum_{j \in L} P_j &= \sum_{j \in H} (P_j - M_j) + \sum_{j \in L} (M_j - P_j)\\
    & = \sum_j |P_j - M_j|\\
    &\geq C\cdot \sqrt{\frac d n}.
\end{align*}

We then consider two functions $f_H$ and $f_L$.
\begin{align*}
    f_H(x)_y = \frac{\mathbf{1}((x,y) \in H)}{\max(1,\sum_y \mathbf{1}((x,y) \in H))} \;\;\;&\;\;\; f_L(x)_y =  \frac{\mathbf{1}((x,y) \in L)}{\max(1,\sum_y \mathbf{1}((x,y) \in L))}
\end{align*}
It is easy to see that given Evelyn's view, the functions $f_H$ and $f_L$ are indistinguishable, since $H$ and $L$ have the same distribution. However the score of $f_H$ and $f_L$ on the true distribution differ by at least $\Omega(\sqrt{\frac {k\calY} n})$. This implies the claimed lower bound.

\end{document}